\documentclass[sigconf]{acmart}

\usepackage{booktabs} 
\usepackage{graphicx}

\setcopyright{none}

\acmDOI{}

\acmISBN{}

\acmConference[SysML]{SysML Conference}{February 15-16, 2018}{Stanford, California, USA} 
\acmYear{2018}
\copyrightyear{}

\begin{document}
\title{Toward Scalable Verification for Safety-Critical Deep Networks}

\author{Lindsey Kuper}
\affiliation{%
  \institution{Parallel Computing Lab, Intel Labs}
}
\email{lindsey.kuper@intel.com}

\author{Guy Katz}
\affiliation{%
  \institution{Stanford University}
}
\email{guyk@cs.stanford.edu}

\author{Justin Gottschlich}
\affiliation{%
  \institution{Parallel Computing Lab, Intel Labs}
}
\email{justin.gottschlich@intel.com}

\author{Kyle Julian}
\affiliation{%
  \institution{Stanford University}
}
\email{kjulian3@stanford.edu}

\author{Clark Barrett}
\affiliation{%
  \institution{Stanford University}
}
\email{barrett@cs.stanford.edu}

\author{Mykel J. Kochenderfer}
\affiliation{%
  \institution{Stanford University}
}
\email{mykel@stanford.edu}

\renewcommand{\shortauthors}{L. Kuper, G. Katz, J. Gottschlich, K. Julian, C. Barrett, M. Kochenderfer}

\begin{abstract}
The increasing use of deep neural networks for safety-critical applications, such as autonomous driving and flight control, raises concerns about their safety and reliability. Formal verification can address these concerns by guaranteeing that a deep learning system operates as intended, but the state of the art is limited to small systems. In this work-in-progress report we give an overview of our work on mitigating this difficulty, by pursuing two complementary directions: devising scalable verification techniques, and identifying design choices that result in deep learning systems that are more amenable to verification.

\end{abstract}

\maketitle

\section{Introduction}


Machine learning systems, and, in particular, deep neural networks (DNNs), are becoming a widely used and effective means for tackling complex, real-world problems~\cite{goodfellow:2016:dl}. However, a major obstacle to the use of DNNs in safety-critical systems, such as autonomous driving or flight control systems, is the great difficulty in providing formal guarantees about their behavior.

A powerful technique for formal verification of properties of a software artifact is to encode the artifact and the property one wishes to prove about it as a \emph{satisfiability modulo theories} (SMT) formula, and then use an SMT solver to prove that the property holds or find a counterexample showing that it does not.  While it is possible to verify properties of neural networks using SMT solvers, until recently the technique only scaled to toy-sized networks of fewer than ten neurons~\cite{pulina2010}.

Yet, for the practical adoption of SMT-based DNN verification, we must be able to verify properties of DNNs of up to thousands (or more) of neurons. To do this, we advocate a two-pronged approach. First, we propose the development of specialized, efficient SMT solvers that are well-suited for DNN verification problems. Second, we propose designing DNNs in ways that make them more amenable to SMT-based verification. These two approaches complement each other, and we observe that design choices that make a DNN more amenable to verification are also desirable for other reasons, such as improved speed of inferencing, smaller memory requirements, and reduced power footprint.

\section{Scaling up SMT-based verification of neural networks}

\begin{figure}[t]
\includegraphics[width=\linewidth]{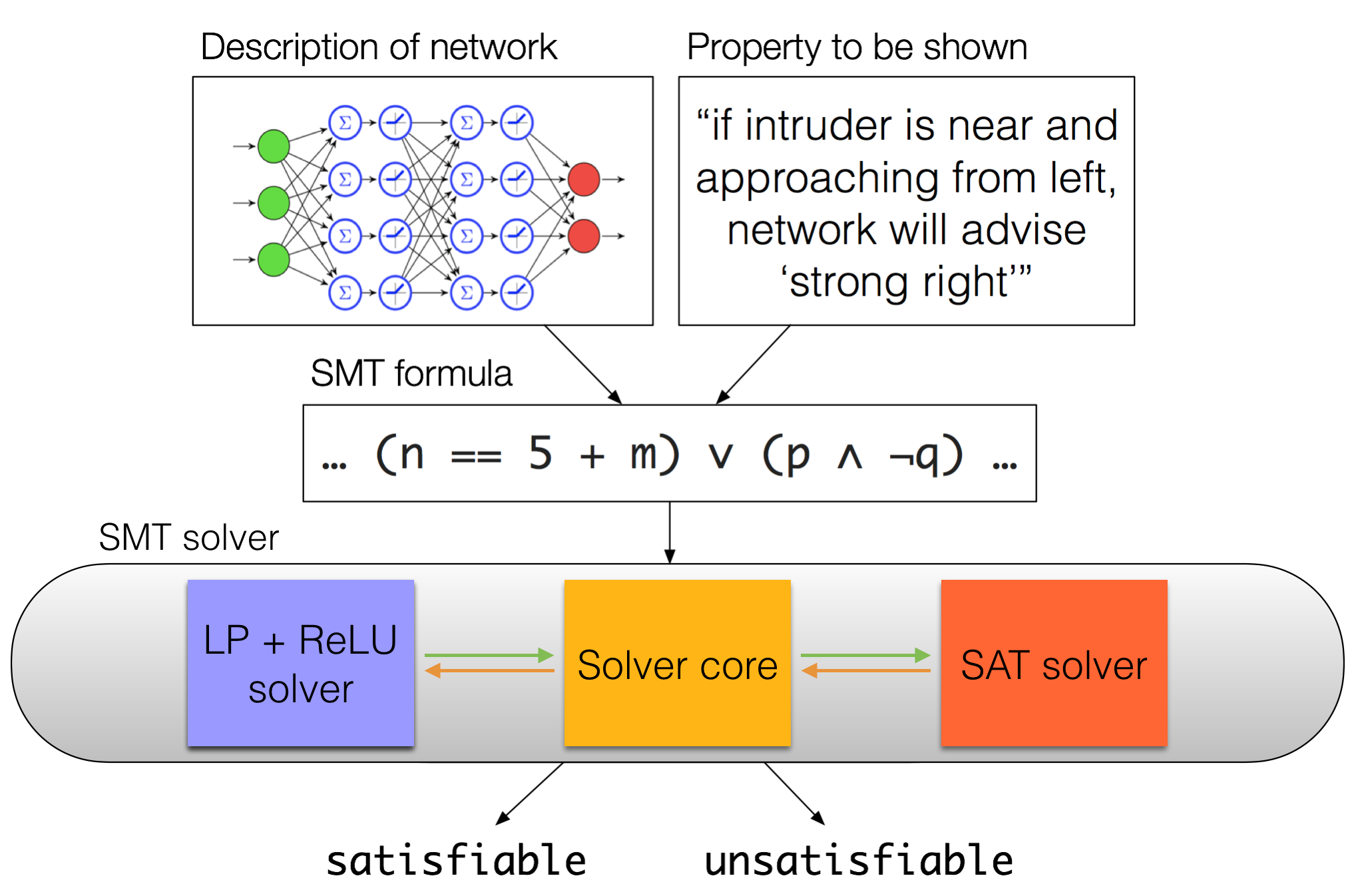}
\caption{\small Overview of the Reluplex architecture.  Reluplex takes as input a network description and a property we wish to prove about the network's behavior, both expressed as an SMT formula.  The SMT solver incorporates a domain-specific linear programming (LP) + ReLU theory solver that interacts with an underlying SAT solver and determines whether the formula is satisfiable.}
\end{figure}

A primary focus of this work is in extending the capabilities of automated verification tools such as SMT solvers to formally verify properties of DNNs used for safety-critical systems. A major challenge of verifying properties of DNNs with SMT solvers is in handling the networks' activation functions. 

Each neuron of a neural network computes a weighted sum of its inputs according to learned weights. It then passes that sum through an activation function to produce the neuron's final output. Typically, the activation functions (e.g., sigmoid) introduce nonlinearity to the network, making DNNs capable of learning arbitrarily complex functions, but also making the job of automated verification tools much harder, in some cases moving the problem from P to NP.

Using SMT solvers to verify properties of neural networks involves encoding the network and the property in question as formulas in some \emph{theory}, such as the theory of linear real arithmetic.  Our work leverages the observation that, apart from their activation functions, neural networks can be expressed using conjunctions of linear real arithmetic formulas, which are straightforward to handle using standard linear programming (LP) solving algorithms.  It is also possible to express \emph{piecewise-linear} activation functions, such as rectified linear units (ReLUs), as part of linear arithmetic formulas, but every ReLU in a network then introduces a disjunction to the formula. These disjunctions quickly cause an exponential increase in the state space that the SMT solver must explore to prove properties about the network, thus limiting the applicability and scalability of the approach.

In a recent paper~\cite{reluplex}, we proposed an improved SMT-based algorithm, called Reluplex, capable of verifying properties of networks that are an order of magnitude larger than previously possible. Reluplex mitigates the difficulty posed by activation functions through a \emph{lazy} approach, which often makes it possible to eliminate many activation functions from the problem without changing the result. It extends the theory of linear real arithmetic by introducing a new ``ReLU'' predicate that can be split into disjuncts lazily, making it possible to avoid exploring large parts of the state space.  As a result, the Reluplex solver can verify networks that are notably larger than what was previously possible. For example, we used Reluplex to verify safety properties of a DNN used as the controller for a prototype of the ACAS Xu aircraft collision avoidance system~\cite{policycompression}.

The lazy-ReLU-splitting technique that Reluplex uses is an example of the general problem-solving strategy of \emph{exploiting high-level domain-specific abstractions for efficiency} that has proven fruitful in a variety of areas.  For example, in the setting of high-performance domain-specific languages, a high-level representation of programmer intent enables compiler optimizations and smart scheduling choices that would be difficult or impossible otherwise~\cite{delite}. The use of high-level abstractions not only does not compromise high performance, but actually enables it.  Reluplex's lazy ReLU splitting is another such optimization, made possible by the addition of the high-level ReLU predicate to the theory used by the solver.  The higher-level representation makes it possible to determine the satisfiability of a formula more efficiently than if the problem were expressed at a lower level.

An important lesson here for scalable verification is that we have much to gain by not treating SMT solvers as black boxes, but instead developing \emph{domain-specific theory solvers} like Reluplex that are uniquely suited to the verification task at hand.  We are currently working on extending Reluplex to handle piecewise-linear approximations of other commonly used activation functions to be able to handle a wider variety of networks.

\section{Designing verification-friendly neural networks}

In addition to improving the scalability of verification tools, a complementary direction for scaling DNN verification is to design the networks themselves in a way that makes them more amenable to verification.  When designing a neural network, some of the obvious design decisions are related to the topology of the network,
such as the number of hidden layers and their dimensions. It is not
surprising that, from a verification point of view, smaller networks
are generally easier to handle. Developers of neural networks may
opt to use a smaller network, perhaps achieving lower accuracy,
in order to enjoy the benefits of verification.  On the other hand, recent work~\cite{deep-compression} suggests that it is possible to significantly reduce the storage requirements of neural networks without compromising accuracy.  Although the motivation for this work was ease of deployment of neural networks in resource-constrained settings rather than ease of verification, these pruned, quantized networks may also be easier for verification tools to handle than uncompressed networks.

Our initial experiments suggest that the size of the network is not necessarily the only factor to consider; the network topology is also important. We have observed that networks with many layers with a few neurons each are generally easier for the solver to handle than networks with few layers, but many neurons in each layer.

An extreme way of applying this principle is by discretizing parts of the neural network in question, effectively turning it into a family of smaller networks.  The ACAS Xu network~\cite{policycompression} used this approach due to considerations that did not include verification --- rather, due to hardware constraints, the developers found that many smaller networks were preferable to one large one. However, the discretization step also made it easier to verify properties of each of the smaller networks. A similar approach could facilitate the verification of other systems as well.

Another decision with consequences for the scalability of verification is activation function selection. These choices can have far-reaching effects. For example, some of the more successful verification efforts thus far~\cite{reluplex, ehlers2017} have focused on piecewise-linear activation functions, such as ReLUs or max-pooling layers, while attempts to verify networks with sigmoid activation functions have proved far less scalable.  

In addition to network topology and activation functions, there are several other potential avenues to explore. One  example is low-precision DNN arithmetic, which is an increasingly popular way to accelerate DNN training and inferencing~\cite{lowprecision}. The simplicity and smaller size of low-precision networks will make them more amenable to verification than full-precision networks~\cite{Narodytska2017,Cheng2017}, as well as more suitable for use on low-power edge devices~\cite{lane:2016:lowpower}. It may even be the case that hardware accelerator techniques that optimize inference on low-precision networks could be used to speed up verification of those same networks.

\section{Conclusion}
Verifying that neural networks behave as intended may soon become a limiting factor in their applicability to real-world, safety-critical systems such as those used to control autonomous vehicles and aircraft.  Recent work revealing neural networks' vulnerability to adversarial inputs~\cite{szegedy2014}, including in physical-world attacks~\cite{adversarial-physical}, makes meeting this challenge more urgent.

Verification is a promising avenue for mitigating this difficulty, but additional work is required to scale up verification techniques to be practically applicable to modern DNNs. Initial work by us and others points to two complementary avenues that could achieve the sought-after scalability: first, the design of verification algorithms tailored for neural networks (e.g., by enriching the theories used by SMT solvers); and second, the creation and use of design principles for neural networks that produce DNNs that are more amenable to verification (e.g., model topology and activation function selection). We believe that through additional work in these directions, verification could be successfully applied to many real-world deep learning systems.

\bibliographystyle{ACM-Reference-Format}
\bibliography{references}


\begin{thebibliography}{13}


\ifx \showCODEN    \undefined \def \showCODEN     #1{\unskip}     \fi
\ifx \showDOI      \undefined \def \showDOI       #1{#1}\fi
\ifx \showISBNx    \undefined \def \showISBNx     #1{\unskip}     \fi
\ifx \showISBNxiii \undefined \def \showISBNxiii  #1{\unskip}     \fi
\ifx \showISSN     \undefined \def \showISSN      #1{\unskip}     \fi
\ifx \showLCCN     \undefined \def \showLCCN      #1{\unskip}     \fi
\ifx \shownote     \undefined \def \shownote      #1{#1}          \fi
\ifx \showarticletitle \undefined \def \showarticletitle #1{#1}   \fi
\ifx \showURL      \undefined \def \showURL       {\relax}        \fi
\providecommand\bibfield[2]{#2}
\providecommand\bibinfo[2]{#2}
\providecommand\natexlab[1]{#1}
\providecommand\showeprint[2][]{arXiv:#2}

\bibitem[\protect\citeauthoryear{Chafi, Sujeeth, Brown, Lee, Atreya, and
  Olukotun}{Chafi et~al\mbox{.}}{2011}]%
        {delite}
\bibfield{author}{\bibinfo{person}{Hassan Chafi}, \bibinfo{person}{Arvind~K.
  Sujeeth}, \bibinfo{person}{Kevin~J. Brown}, \bibinfo{person}{HyoukJoong Lee},
  \bibinfo{person}{Anand~R. Atreya}, {and} \bibinfo{person}{Kunle Olukotun}.}
  \bibinfo{year}{2011}\natexlab{}.
\newblock \showarticletitle{A Domain-Specific Approach to Heterogeneous
  Parallelism}. In \bibinfo{booktitle}{\emph{Proceedings of the 16th ACM
  Symposium on Principles and Practice of Parallel Programming}}
  \emph{(\bibinfo{series}{PPoPP '11})}. \bibinfo{publisher}{ACM},
  \bibinfo{address}{New York, NY, USA}, \bibinfo{pages}{35--46}.
\newblock
\showISBNx{978-1-4503-0119-0}
\urldef\tempurl%
\url{https://doi.org/10.1145/1941553.1941561}
\showDOI{\tempurl}


\bibitem[\protect\citeauthoryear{Cheng, N{\"{u}}hrenberg, and Ruess}{Cheng
  et~al\mbox{.}}{2017}]%
        {Cheng2017}
\bibfield{author}{\bibinfo{person}{Chih{-}Hong Cheng}, \bibinfo{person}{Georg
  N{\"{u}}hrenberg}, {and} \bibinfo{person}{Harald Ruess}.}
  \bibinfo{year}{2017}\natexlab{}.
\newblock \showarticletitle{Verification of Binarized Neural Networks}.
\newblock \bibinfo{journal}{\emph{CoRR}}  \bibinfo{volume}{abs/1710.03107}
  (\bibinfo{year}{2017}).
\newblock
\showeprint[arxiv]{1710.03107}
\urldef\tempurl%
\url{http://arxiv.org/abs/1710.03107}
\showURL{%
\tempurl}


\bibitem[\protect\citeauthoryear{Ehlers}{Ehlers}{2017}]%
        {ehlers2017}
\bibfield{author}{\bibinfo{person}{R{\"{u}}diger Ehlers}.}
  \bibinfo{year}{2017}\natexlab{}.
\newblock \showarticletitle{Formal Verification of Piece-Wise Linear
  Feed-Forward Neural Networks}.
\newblock \bibinfo{journal}{\emph{CoRR}}  \bibinfo{volume}{abs/1705.01320}
  (\bibinfo{year}{2017}).
\newblock
\showeprint[arxiv]{1705.01320}
\urldef\tempurl%
\url{http://arxiv.org/abs/1705.01320}
\showURL{%
\tempurl}


\bibitem[\protect\citeauthoryear{Goodfellow, Bengio, and Courville}{Goodfellow
  et~al\mbox{.}}{2016}]%
        {goodfellow:2016:dl}
\bibfield{author}{\bibinfo{person}{Ian Goodfellow}, \bibinfo{person}{Yoshua
  Bengio}, {and} \bibinfo{person}{Aaron Courville}.}
  \bibinfo{year}{2016}\natexlab{}.
\newblock \bibinfo{booktitle}{\emph{Deep Learning}}.
\newblock \bibinfo{publisher}{The MIT Press}.
\newblock
\showISBNx{0262035618, 9780262035613}
\newblock
\shownote{\url{http://www.deeplearningbook.org}.}


\bibitem[\protect\citeauthoryear{Han, Mao, and Dally}{Han
  et~al\mbox{.}}{2015}]%
        {deep-compression}
\bibfield{author}{\bibinfo{person}{Song Han}, \bibinfo{person}{Huizi Mao},
  {and} \bibinfo{person}{William~J. Dally}.} \bibinfo{year}{2015}\natexlab{}.
\newblock \showarticletitle{Deep Compression: Compressing Deep Neural Network
  with Pruning, Trained Quantization and Huffman Coding}.
\newblock \bibinfo{journal}{\emph{CoRR}}  \bibinfo{volume}{abs/1510.00149}
  (\bibinfo{year}{2015}).
\newblock
\showeprint[arxiv]{1510.00149}
\urldef\tempurl%
\url{http://arxiv.org/abs/1510.00149}
\showURL{%
\tempurl}


\bibitem[\protect\citeauthoryear{Hubara, Courbariaux, Soudry, El{-}Yaniv, and
  Bengio}{Hubara et~al\mbox{.}}{2016}]%
        {lowprecision}
\bibfield{author}{\bibinfo{person}{Itay Hubara}, \bibinfo{person}{Matthieu
  Courbariaux}, \bibinfo{person}{Daniel Soudry}, \bibinfo{person}{Ran
  El{-}Yaniv}, {and} \bibinfo{person}{Yoshua Bengio}.}
  \bibinfo{year}{2016}\natexlab{}.
\newblock \showarticletitle{Quantized Neural Networks: Training Neural Networks
  with Low Precision Weights and Activations}.
\newblock \bibinfo{journal}{\emph{CoRR}}  \bibinfo{volume}{abs/1609.07061}
  (\bibinfo{year}{2016}).
\newblock
\showeprint[arxiv]{1609.07061}
\urldef\tempurl%
\url{http://arxiv.org/abs/1609.07061}
\showURL{%
\tempurl}


\bibitem[\protect\citeauthoryear{Julian, Lopez, Brush, Owen, and
  Kochenderfer}{Julian et~al\mbox{.}}{2016}]%
        {policycompression}
\bibfield{author}{\bibinfo{person}{Kyle Julian}, \bibinfo{person}{Jessica
  Lopez}, \bibinfo{person}{Jeffrey~S. Brush}, \bibinfo{person}{Michael Owen},
  {and} \bibinfo{person}{Mykel~J. Kochenderfer}.}
  \bibinfo{year}{2016}\natexlab{}.
\newblock \showarticletitle{Policy Compression for Aircraft Collision Avoidance
  Systems}. In \bibinfo{booktitle}{\emph{Digital Avionics Systems Conference
  (DASC)}}.
\newblock
\urldef\tempurl%
\url{https://doi.org/10.1109/DASC.2016.7778091}
\showDOI{\tempurl}


\bibitem[\protect\citeauthoryear{Katz, Barrett, Dill, Julian, and
  Kochenderfer}{Katz et~al\mbox{.}}{2017}]%
        {reluplex}
\bibfield{author}{\bibinfo{person}{Guy Katz}, \bibinfo{person}{Clark~W.
  Barrett}, \bibinfo{person}{David~L. Dill}, \bibinfo{person}{Kyle Julian},
  {and} \bibinfo{person}{Mykel~J. Kochenderfer}.}
  \bibinfo{year}{2017}\natexlab{}.
\newblock \showarticletitle{Reluplex: An Efficient {SMT} Solver for Verifying
  Deep Neural Networks}. In \bibinfo{booktitle}{\emph{Computer Aided
  Verification - 29th International Conference, {CAV} 2017, Heidelberg,
  Germany, July 24-28, 2017, Proceedings, Part {I}}}. \bibinfo{pages}{97--117}.
\newblock
\urldef\tempurl%
\url{https://doi.org/10.1007/978-3-319-63387-9_5}
\showDOI{\tempurl}


\bibitem[\protect\citeauthoryear{Kurakin, Goodfellow, and Bengio}{Kurakin
  et~al\mbox{.}}{2016}]%
        {adversarial-physical}
\bibfield{author}{\bibinfo{person}{Alexey Kurakin}, \bibinfo{person}{Ian~J.
  Goodfellow}, {and} \bibinfo{person}{Samy Bengio}.}
  \bibinfo{year}{2016}\natexlab{}.
\newblock \showarticletitle{Adversarial examples in the physical world}.
\newblock \bibinfo{journal}{\emph{CoRR}}  \bibinfo{volume}{abs/1607.02533}
  (\bibinfo{year}{2016}).
\newblock
\showeprint[arxiv]{1607.02533}
\urldef\tempurl%
\url{http://arxiv.org/abs/1607.02533}
\showURL{%
\tempurl}


\bibitem[\protect\citeauthoryear{Lane, Bhattacharya, Georgiev, Forlivesi, Jiao,
  Qendro, and Kawsar}{Lane et~al\mbox{.}}{2016}]%
        {lane:2016:lowpower}
\bibfield{author}{\bibinfo{person}{N.~D. Lane}, \bibinfo{person}{S.
  Bhattacharya}, \bibinfo{person}{P. Georgiev}, \bibinfo{person}{C. Forlivesi},
  \bibinfo{person}{L. Jiao}, \bibinfo{person}{L. Qendro}, {and}
  \bibinfo{person}{F. Kawsar}.} \bibinfo{year}{2016}\natexlab{}.
\newblock \showarticletitle{DeepX: A Software Accelerator for Low-Power Deep
  Learning Inference on Mobile Devices}. In \bibinfo{booktitle}{\emph{2016 15th
  ACM/IEEE International Conference on Information Processing in Sensor
  Networks (IPSN)}}. \bibinfo{pages}{1--12}.
\newblock
\urldef\tempurl%
\url{https://doi.org/10.1109/IPSN.2016.7460664}
\showDOI{\tempurl}


\bibitem[\protect\citeauthoryear{Narodytska, Kasiviswanathan, Ryzhyk, Sagiv,
  and Walsh}{Narodytska et~al\mbox{.}}{2017}]%
        {Narodytska2017}
\bibfield{author}{\bibinfo{person}{N. Narodytska}, \bibinfo{person}{S.~P.
  Kasiviswanathan}, \bibinfo{person}{L. Ryzhyk}, \bibinfo{person}{M. Sagiv},
  {and} \bibinfo{person}{T. Walsh}.} \bibinfo{year}{2017}\natexlab{}.
\newblock \showarticletitle{Verifying Properties of Binarized Deep Neural
  Networks}.
\newblock \bibinfo{journal}{\emph{CoRR}}  \bibinfo{volume}{abs/1709.06662}
  (\bibinfo{year}{2017}).
\newblock
\showeprint[arxiv]{1709.06662}
\urldef\tempurl%
\url{http://arxiv.org/abs/1709.06662}
\showURL{%
\tempurl}


\bibitem[\protect\citeauthoryear{Pulina and Tacchella}{Pulina and
  Tacchella}{2010}]%
        {pulina2010}
\bibfield{author}{\bibinfo{person}{Luca Pulina} {and} \bibinfo{person}{Armando
  Tacchella}.} \bibinfo{year}{2010}\natexlab{}.
\newblock \showarticletitle{An Abstraction-refinement Approach to Verification
  of Artificial Neural Networks}. In \bibinfo{booktitle}{\emph{Proceedings of
  the 22nd International Conference on Computer Aided Verification}}
  \emph{(\bibinfo{series}{CAV'10})}. \bibinfo{publisher}{Springer-Verlag},
  \bibinfo{address}{Berlin, Heidelberg}, \bibinfo{pages}{243--257}.
\newblock
\showISBNx{3-642-14294-X, 978-3-642-14294-9}
\urldef\tempurl%
\url{https://doi.org/10.1007/978-3-642-14295-6_24}
\showDOI{\tempurl}


\bibitem[\protect\citeauthoryear{Szegedy, Zaremba, Sutskever, Bruna, Erhan,
  Goodfellow, and Fergus}{Szegedy et~al\mbox{.}}{2013}]%
        {szegedy2014}
\bibfield{author}{\bibinfo{person}{Christian Szegedy},
  \bibinfo{person}{Wojciech Zaremba}, \bibinfo{person}{Ilya Sutskever},
  \bibinfo{person}{Joan Bruna}, \bibinfo{person}{Dumitru Erhan},
  \bibinfo{person}{Ian~J. Goodfellow}, {and} \bibinfo{person}{Rob Fergus}.}
  \bibinfo{year}{2013}\natexlab{}.
\newblock \showarticletitle{Intriguing properties of neural networks}.
\newblock \bibinfo{journal}{\emph{CoRR}}  \bibinfo{volume}{abs/1312.6199}
  (\bibinfo{year}{2013}).
\newblock
\showeprint[arxiv]{1312.6199}
\urldef\tempurl%
\url{http://arxiv.org/abs/1312.6199}
\showURL{%
\tempurl}


\end{thebibliography}

\end{document}